\documentclass[10pt,twocolumn,letterpaper]{article}

\usepackage{iccv}

\usepackage{array}
\usepackage{amsmath}
\usepackage{amssymb}
\usepackage{times}
\usepackage{epsfig}
\usepackage{graphicx}
\usepackage{amsmath}
\usepackage{amssymb}
\usepackage{algorithm}
\usepackage[noend]{algpseudocode}
\usepackage{multirow}
\usepackage[tight,normalsize,sf,SF]{subfigure}
\usepackage{footnote}
\usepackage{soul}

\usepackage{ulem} 
\usepackage[bottom]{footmisc}

\usepackage[pagebackref=true,breaklinks=true,letterpaper=true,colorlinks,bookmarks=false]{hyperref}

\iccvfinalcopy 

\def\iccvPaperID{****} 

\begin{document}

\title{Learn to Scale: Generating Multipolar Normalized Density Maps for Crowd Counting}





\author{Chenfeng Xu$^{1}$\thanks{This work was done when Chenfeng Xu was a research intern at Microsoft Research Asia.}, Kai Qiu$^{2}$, Jianlong Fu$^{2}$, Song Bai$^3$, Yongchao Xu$^1$\thanks{Corresponding author}, Xiang Bai$^1$\\
$^1${Huazhong University of Science and Technology}, $^2${Microsoft Research Asia}, $^3${University of Oxford}\\ 
{\tt \small \{xuchenfeng, yongchaoxu, xbai\}@hust.edu.cn, } {\tt \small \{kaqiu, jianf\}@microsoft.com, } {\tt \small songbai.site@gmail.com}
}

\maketitle


\begin{abstract}
Dense crowd counting aims to predict thousands of human instances from an image, by calculating integrals of a density map over image pixels. Existing approaches mainly suffer from the extreme density variances. Such density pattern shift poses challenges even for multi-scale model ensembling. In this paper, we propose a simple yet effective approach to tackle this problem. First, a patch-level density map is extracted by a density estimation model and further grouped into several density levels which are determined over full datasets. Second, each patch density map is automatically normalized by an online center learning strategy with a multipolar center loss.
Such a design can significantly condense the density distribution into several clusters, and enable that the density variance can be learned by a single model. Extensive experiments demonstrate the superiority of the proposed method. Our work outperforms the state-of-the-art by 4.2\%, 14.3\%, 27.1\% and 20.1\% in MAE, on ShanghaiTech Part A, ShanghaiTech Part B, UCF\_CC\_50 and UCF-QNRF datasets, respectively.

\end{abstract}

\section{Introduction}
\label{section:intro}
A robust crowd counting system is of significantly value in many real-world applications such as video surveillance, security alerting, event planning, \textit{etc}. In recent years, the deep learning based approaches have been the mainstream of crowd counting, due to the powerful representation learning ability of convolutional neural networks (CNNs). To estimate the count, predominant approaches generate a density map by CNN, from which the count of instances can be integrated over image pixels.

\begin{figure}[!tb]
\centering
\subfigure[]
{
\begin{minipage}[tb]{0.475\textwidth}
\includegraphics[width=1\linewidth]{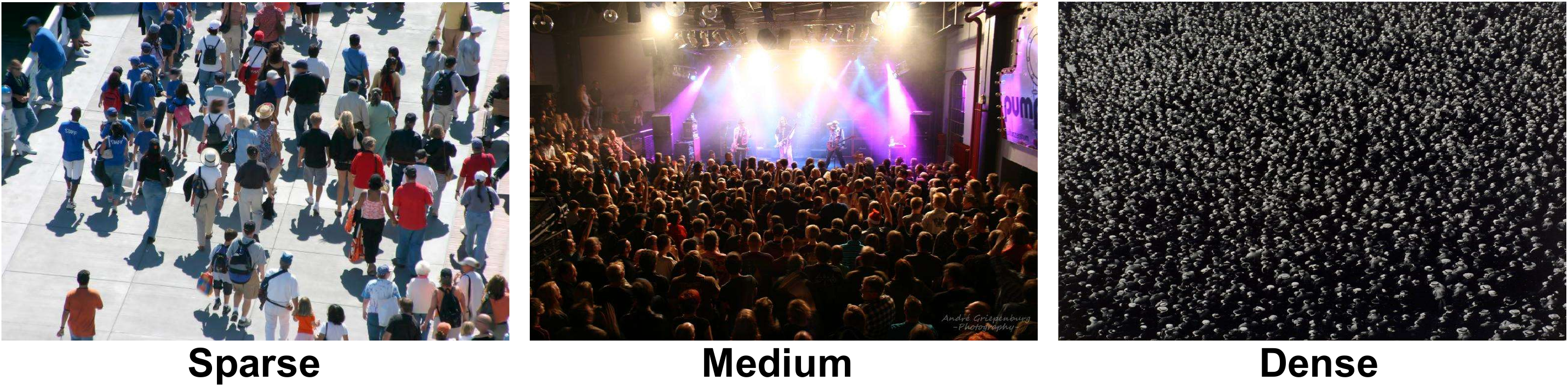}
\end{minipage}
\label{fig:song_1}
}
\subfigure[]
{
\begin{minipage}[tb]{0.475\textwidth}
\includegraphics[width=1\linewidth]{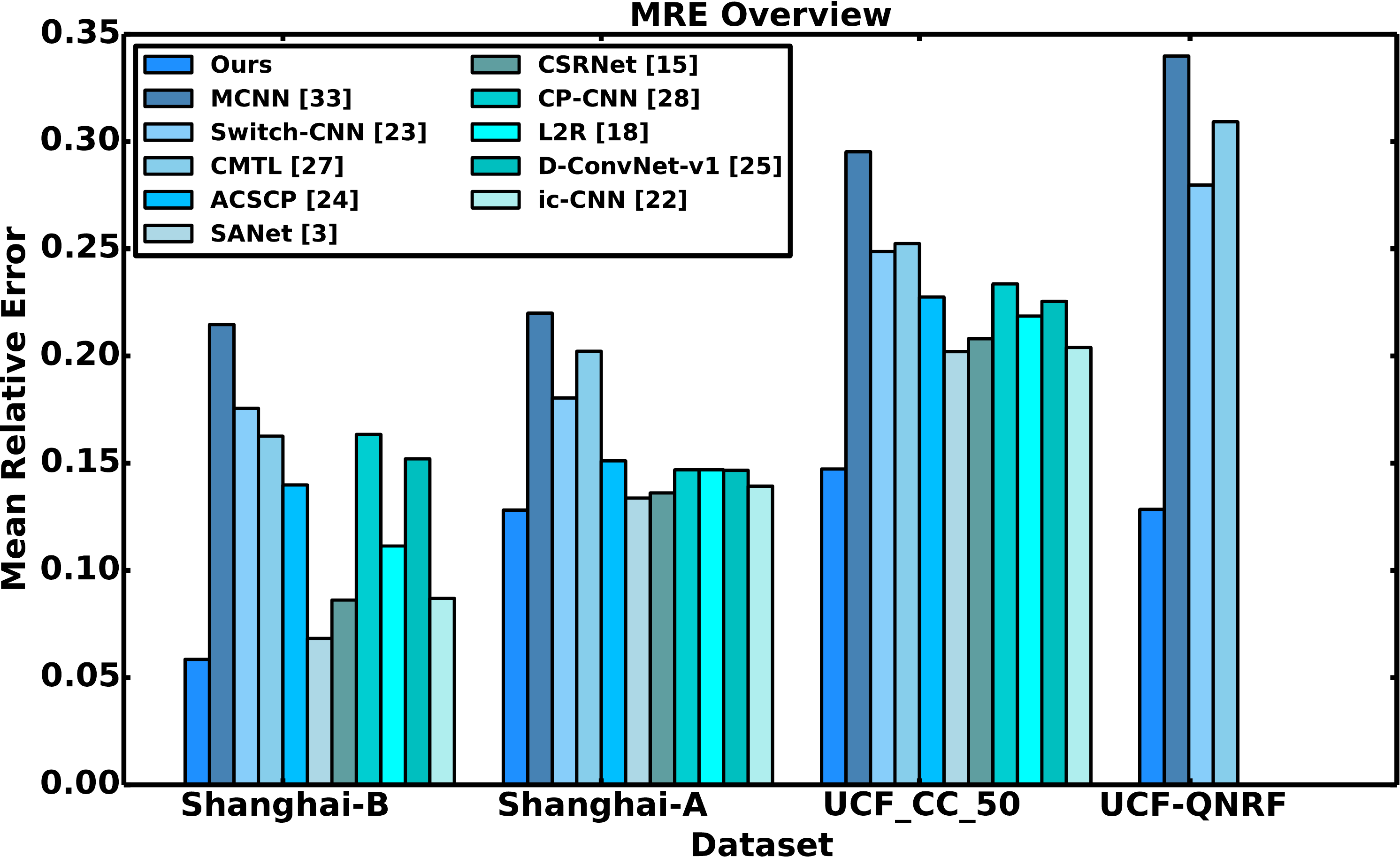}
\end{minipage}
\label{fig:song_2}
}
\caption{(a) Three examples from ShanghaiTech Part A dataset, which show extreme density variances. (b) Comparison of Mean Relative Error on four crowd counting datasets (the scale variances get larger from left to right) of different approaches. Results show the robustness of the proposed approach to extreme scale variances. [best viewed in color].}
\label{fig:distribution map}
\end{figure}

Although crowd counting has been extensively studied by previous methods, handling the large density variances which cause huge density pattern shift in crowd images is still an open issue. As illustrated in Fig.~\ref{fig:song_1}, the densities of crowd image patches can vary significantly, which change from a bit sparse (\textit{e.g.}, ShanghaiTech Part B) to extremely dense (\textit{e.g.}, UCF-QNRF). Such large density pattern shifts usually bring grand challenges to density prediction by a single CNN model, due to its fixed sizes of receptive fields.
Remarkable progress has been achieved by learning a density map through designing multi-scale architectures~\cite{sam2017switching} or aggregating multi-scale features~\cite{cao2018scale,zhang2016single}, which indicate that the ability to cope with density variation is crucial for crowd counting methods. Although density maps with multiple scales can be generated and aggregated, it is still hard to ensure robustness when the density variances get increased a lot. As shown in Fig.~\ref{fig:song_2}, most recent works obtain a higher MRE\footnote{MRE is calculated by MAE/P, where MAE denotes the standard Mean Average Error and P is the average count of a dataset} on datasets with larger density variances, which indicates that the extreme density variance and pattern shift in crowd counting remains a huge challenge.

In this paper, we propose a simple yet effective method to mitigate the problem caused by extreme density variances.
The core idea is learning to scale image patches and to facilitate the density distribution condensing to several clusters, thus the density variance can be reduced. The scale factor of each image patch can be automatically learned during training, with the supervision of a novel multipolar center loss (MPCL).
More specifically, all the patches at each density level are optimized to approach a density center, which can be updated by online calculating a mean value for each density level.

In particular, the proposed framework consists of two closely-related steps.
First, given an image, an initial density map is generated by our designed Scale Preserving Network (SPN). After that, each density map is divided into $K \times K$ patches, and all the patch-level density maps are further evenly divided into $G$ groups, according to their density levels.
Second, each patch is scaled by a learned scale factor, thus the density of this patch can converge to a center of its density level. The final density map for the input image can be obtained by concatenating the $K \times K$ patch-level density maps.

Experiments are conducted on several popular benchmark datasets, including ShanghaiTech~\cite{zhang2016single}, UCF\_CC\_50~\cite{idrees2013multi}, and UCF-QNRF~\cite{idrees2018composition}. Extensive evaluations demonstrate superior performance over the prior arts. Moreover, the cross dataset validation on these datasets further demonstrates that the proposed method has a powerful transferability. In summary, the main contributions in this paper are two-fold:
\begin{itemize}
\item[-] We propose a Learning to Scale Module (L2SM) to solve the density variation issue in crowd counting. With L2SM, different regions can be automatically scaled so that they have similar densities, while the quality of the density maps is significantly improved. L2SM is end-to-end trainable when adding it into a CNN model for density estimation.

\item[-] The proposed L2SM added into SPN significantly outperforms state-of-the-art methods on three widely-adopted challenging datasets, demonstrating its effectiveness in handling density variation. Furthermore, L2SM also has a good transferability under cross dataset validation on different datasets, showing the generalizability of the proposed method. 
 
\end{itemize}

\section{Related Work}
\label{sec:relatedwork}

Crowd counting has attracted much attention in computer vision. Early methods frame the counting problem as a detection task~\cite{5975165,viola2003detecting} that explicitly detects individual heads, which has major difficulty in occlusion and dense areas. The regression-based methods~\cite{chan2008privacy,chen2012feature,ge2009marked,idrees2015detecting} greatly improve the counting performance on dense areas via different regression functions such as Gaussian process, ridge regression, and random forest regression. Recently, with the development of deep learning, the mainstream crowd counting methods switch to CNN-based methods~\cite{pham2015count,zhang2015cross,boominathan2016crowdnet,zhang2016single,xiong2017spatiotemporal,chattopadhyay2017counting,liu2018leveraging}. These CNN-based methods address the crowd counting via regressing density map representations~\cite{lempitsky2010learning}, and achieve higher accuracy and transferability than the classical methods. Recent methods mainly focus on two challenging aspects faced by current CNN-based methods: huge scale and density variance and severe over-fitting.


\textbf{Methods addressing huge scale and density variance.} Multi-scale is a challenging problem for many vision tasks including crowd counting. It is difficult to count the small heads in dense areas accurately. There are many methods attempting to handle huge scale variance. The existing methods can be roughly divided into two categories: methods that explicitly rely on scale information and methods that implicitly cope with multi-scale.  


1) Some methods explicitly make use of scale information for crowd counting. For instance, Zhang~\textit{et al.}~\cite{zhang2015cross} and Onoro~\textit{et al.}~\cite{onoro2016towards} adopt CNNs with provided geometric or perspective information. Yet, this scale related information is not always readily available. Sindagi~\textit{et al.}~\cite{sindagi2017generating} use networks to estimate the density degree for the corresponding whole and partial region based on manually setting scale degrees and fuse them as context information. Sam~\textit{et al.}~\cite{sam2017switching} leverage the scale information to design different networks for dividing and counting.
To overcome the difficulty in manually setting the scale degree, Sam~\textit{et al.}~\cite{babu2018divide} design an incrementally growing CNN to deal with areas of different density degrees without involving any handcraft steps.

2) Some other works aim to implicitly cope with the multi-scale problem. Zhang~\textit{et al.}~\cite{zhang2016single} and Cao~\textit{et al.}~\cite{cao2018scale} propose to build a multi-column CNN to extract multi-scale features and fuse them together for density map estimation. Different from multi-scale feature fusion, Liu~\textit{et al.}~\cite{liu2019context} attempt to encode the scale of the contextual information required to predict crowd density accurately. In~\cite{li2018csrnet}, Li~\textit{et al.} propose to increment the receptive field size in CNN to better leverage multi-scale information. In addition to these specific network designs for implicitly handling the multi-scale problem, Shen \textit{et al.}~\cite{shen2018crowd} introduce an~\textit{ad hoc} term in the training loss function in order to pursue the cross-scale consistency. In~\cite{idrees2018composition}, Idrees~\textit{et al.} propose to adopt variant ground-truth density map representation with Gaussian kernels of different sizes to better deal with density map estimation in areas of different density levels.

\textbf{Methods alleviating severe over-fitting.} It is well-known that deep CNNs~\cite{lecun2015deep,vgg16network} usually struggle with the over-fitting problem on small datasets. Current CNN-based crowd counting methods also face this challenge due to the small size and limited variety of existing datasets, leading to weak performance and transferability. To overcome the over-fitting, Liu~\textit{et al.}~\cite{liu2018leveraging} propose a learning-to-rank framework to leverage abundantly available unlabeled crowd images and a self-learning strategy. Shi~\textit{et al.}~\cite{shi2018crowd} build a set of decorrelated regressors with reasonable generalization capabilities through managing their intrinsic diversities to avoid severe over-fitting.

Though many methods have been proposed to tackle the large scale and density variation issue, this problem still remains challenging for crowd counting. Different from previous methods~\cite{zhang2016single,sam2017switching,sindagi2017cnn,babu2018divide,cao2018scale, liu2018crowd}, we mimic a rational human behavior in crowd counting through learning to scale dense region counting. We compute the scale ratios with a novel use of multipolar center loss~\cite{wen2016discriminative} to explicitly bring all the regions of significantly varied density to multiple similar density levels. This results in a robust density estimation on dense regions and appealing transferability.



\begin{figure}
\centering
\includegraphics[width=1\linewidth]{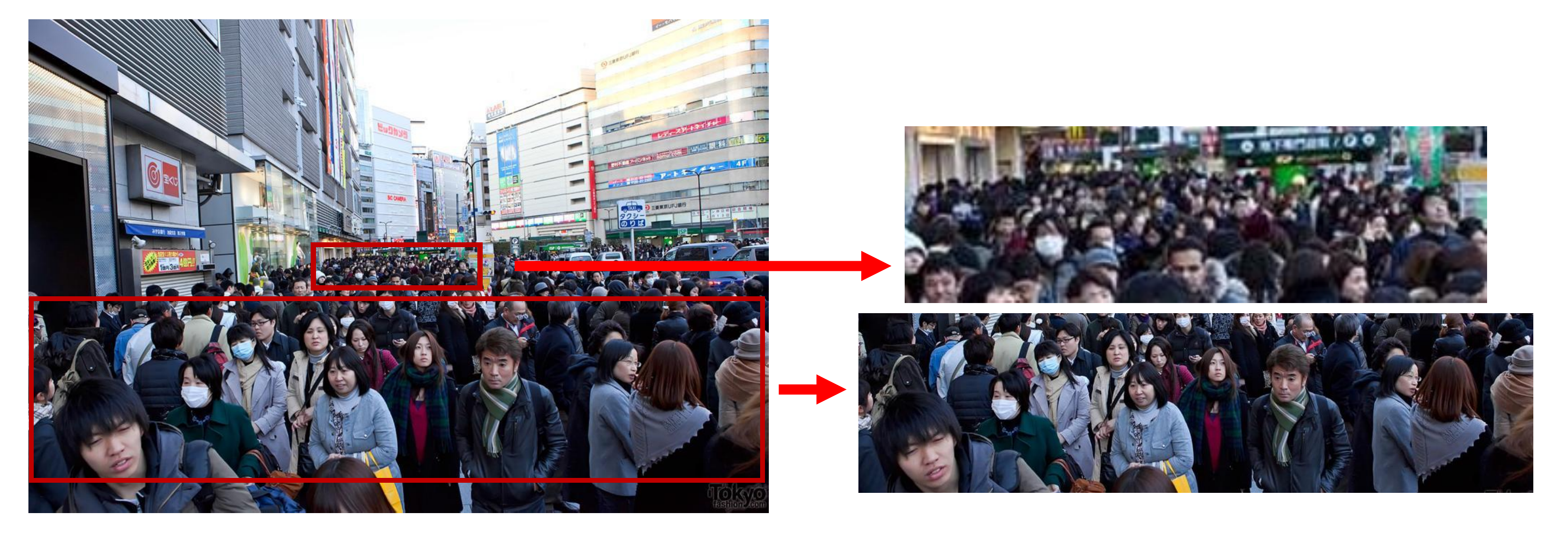}
\caption{A rational human behavior. For a given image, we are prone to first count in the regions of large heads (\textit{e.g.}, region on the bottom of image), then zoom in the regions of dense small heads for precise counting (see for example the region in the middle and its zoomed version on top right).
}
\label{fig:img_example}
\end{figure}

\begin{figure*}[ht]
\centering
\includegraphics[width=0.75\paperwidth]{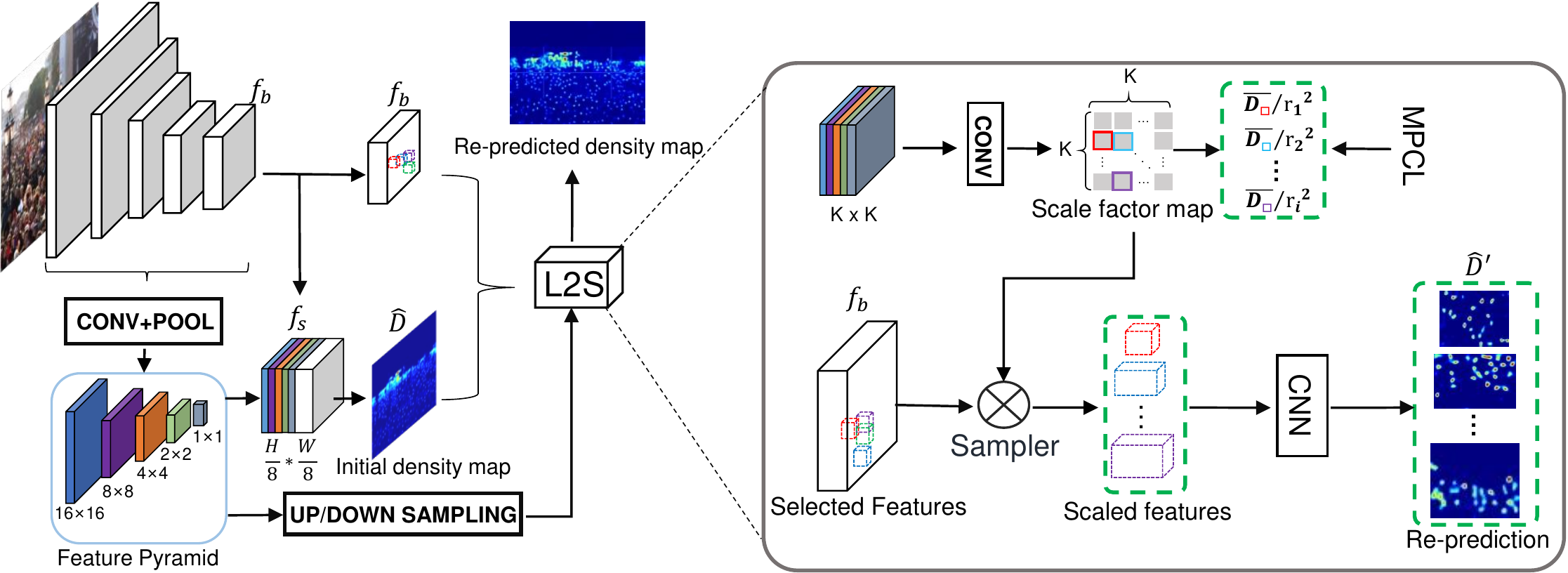}
\centering
\caption{Overall pipeline of the proposed method with two modules: 1) Scale Preserving Network (SPN) to generate an initial density map $\hat{D}$ from stacked feature $f_s$, and 2) Learning to Scale Module (L2SM) that computes the scale ratios $r$ for dense regions selected (based on $\hat{D}$) from $K \times K$ non-overlapping divisions of image domain, and then re-predicts the density map $\hat{D}'$ for selected dense regions from scaled feature $f_b$. We adopt multipolar center loss (MPCL) on relative density level reflected by $\hat{D}_i/r_i^2$ for each region $R_i$ to explicitly centralize all the selected dense regions into multiple similar density levels. This alleviates the density pattern shift issue caused by the large density variation between sparse and dense regions.}
\label{fig:architecture}
\end{figure*}


\section{Method}

\subsection{Overview}
\label{subsec:overview}
The mainstream crowd counting methods model the problem as density map regression using CNNs. For a given image, the ground-truth density map $D$ is given by spreading binary head locations to nearby regions with Gaussian kernels. For sparse regions, the ground-truth density only depends on a specific person, resulting in regular Gaussian blobs. For dense regions, multiple crowded heads may spread to the same nearby pixel, yielding high ground-truth densities with very different density patterns compared with sparse regions. These density pattern shifts make it difficult to accurately predict the density maps for both dense and sparse regions in the same way. 



To improve the counting accuracy, we aim to tackle the problem of pattern shift caused by large density variations and refine the prediction for highly dense regions. Specifically, the proposed method mimics a rational behavior when humans count crowds. For a given crowd image, we are prone to begin with dividing the image into partitions of different crowding levels before attempting to count the people. For sparse regions of large heads, it is easy to count the people on the original region directly. Whereas, for dense regions composed of crowded small heads, we need to zoom in the region for more accurate counting. An example of this counting behavior is depicted in Fig.~\ref{fig:img_example}.

We propose a network to mimic such human behavior for crowd counting. The overall pipeline is depicted in Fig.~\ref{fig:architecture}, consisting of two modules: 1) Scale preserving network (SPN) presented in Sec.~\ref{section:cdm}. We leverage multi-scale feature fusion to generate an initial density map prediction, which provides an accurate prediction on sparse regions and indicates the density distribution over the image; 2) Learning to scale module (L2SM) detailed in Sec.~\ref{subsec:DCN}. We divide the image into $K \times K$ non-overlapping regions, and select some dense regions (based on the initial density estimation) to re-predict the density map. Specifically, we leverage SPN to compute a scaling factor for each selected dense region, and scale the ground-truth density map by changing the distance between blobs and keeping the same peaks. The density re-prediction for the selected regions is then performed on the scaled features. The key to this re-prediction process lies in computing appropriate scaling factors. For that, we adopt the center loss to centralize the density distributions into multipolar centers, alleviating the density pattern shift issue and thus improving the prediction accuracy. The whole network is end-to-end trainable and the training objective is presented in Sec.~\ref{lossfunc}.




\subsection{Scale Preserving Network}
\label{section:cdm}

We follow the mainstream crowd counting methods by regressing density maps. Precisely, we use geometry-adaptive kernels to generate ground-truth density maps in highly congested scenes. For a given image containing $P$ person, the ground-truth annotation can be represented via a delta function on each pixel $p$: $H(p) = \sum_{i=1}^P{\delta(p - p_i)}$,
where $p_i$ is the annotated location of $i$-th person.
The density map $D$ on each pixel $p$ is then generated by convolving $H(p)$ with a Gaussian kernel $G$: $D(p) = \sum_{i=1}^P{\delta(p - p_i)} * G_{\sigma_i}$,
where the Gaussian kernel $\sigma_i$ is a spread parameter.

We develop a CNN to regress the density map $D$. For a fair comparison with most methods, we adopt VGG16~\cite{vgg16network} as the backbone network. We discard the pooling layer between \textit{stage4} and \textit{stage5}, as well as the last pooling layer and the fully connected layers that follow to preserve accurate spatial information. It is well-known that deep layers in CNN encode more semantic and high-level information, and shallow layers provide more precise localization information. We extract features from different stages by applying $3 \times 3$ convolutions on the last layer of each stage. Then we pool these features extracted from \textit{stage1} to \textit{stage5} into $16 \times 16$, $8 \times 8$, $4 \times 4$, $2 \times 2$, and $1 \times 1$, respectively. This results in a pyramid structure.
Each spatial unit in the pooled feature indicates the density level, hence it maps to the scale information of the underlying image. These scale preserving features are then upsampled to the size of \textit{conv5} by bilinear interpolation and stacked together with features in \textit{conv5} $f_b$. We then feed the stacked feature $f_s$ to three successive convolutions and one deconvolution layer for regressing the density map $\hat{D}$.





\subsection{Learning to Scale Module}
\label{subsec:DCN}
The initial density prediction is accurate on sparse regions thanks to the regular individual Gaussian blobs, but the prediction is less accurate on dense regions composed of crowded heads lying very close to each other. As indicated in Sec.~\ref{subsec:overview}, this triggers the pattern shift on the target density map. Following the rational human behavior in crowd counting, we zoom in the dense regions for better counting accuracy. In fact, on the zoomed version, the distance between nearby heads is enlarged, which results in regular individual Gaussian blobs of target density map, alleviating the density pattern shift. Such density pattern modulating facilitates the prediction. Inspired by this, we first evenly divide the image domain into $K \times K$ (\textit{e.g.}, $K = 4$) non-overlapping regions. We then select the dense regions based on the average initial density $\overline{D}_i = \sum_{p \in R_i} \hat{D}(p)/|R_i|$ of each region $R_i$, where $|R_i|$ denotes the area of region $R_i$.


We achieve this by learning to scale the selected dense regions.

We first leverage the scale preserving pyramid features described in Sec.~\ref{section:cdm} to compute the scaling ratio $r_i$ for each selected region $R_i$.
Precisely, we downsample/upsample the pooled features described in Sec.~\ref{section:cdm} to $K \times K$, and concatenate them together. This is followed by a $1 \times 1$ convolution to produce the scale factor map $r$. Each value in this $K \times K$ map $r$ represents the scaling ratio for the underlying region.

Once having the scale factor map $r$, we scale the feature $f_b$ on the selected regions accordingly through bilinear upsampling.
Based on the scaled feature map corresponding to each selected region $R_{i}$, we apply five successive convolutions to re-predict the density map for scaled $R_i$. 
We then resize the re-predicted density map to the original size of $R_i$ and multiply the density on each pixel by $r_i^2$ to preserve the same counting result. The initial prediction on selected regions is replaced by the re-prediction of resized density map.  

\begin{figure}
\centering
\includegraphics[width=0.8\linewidth]{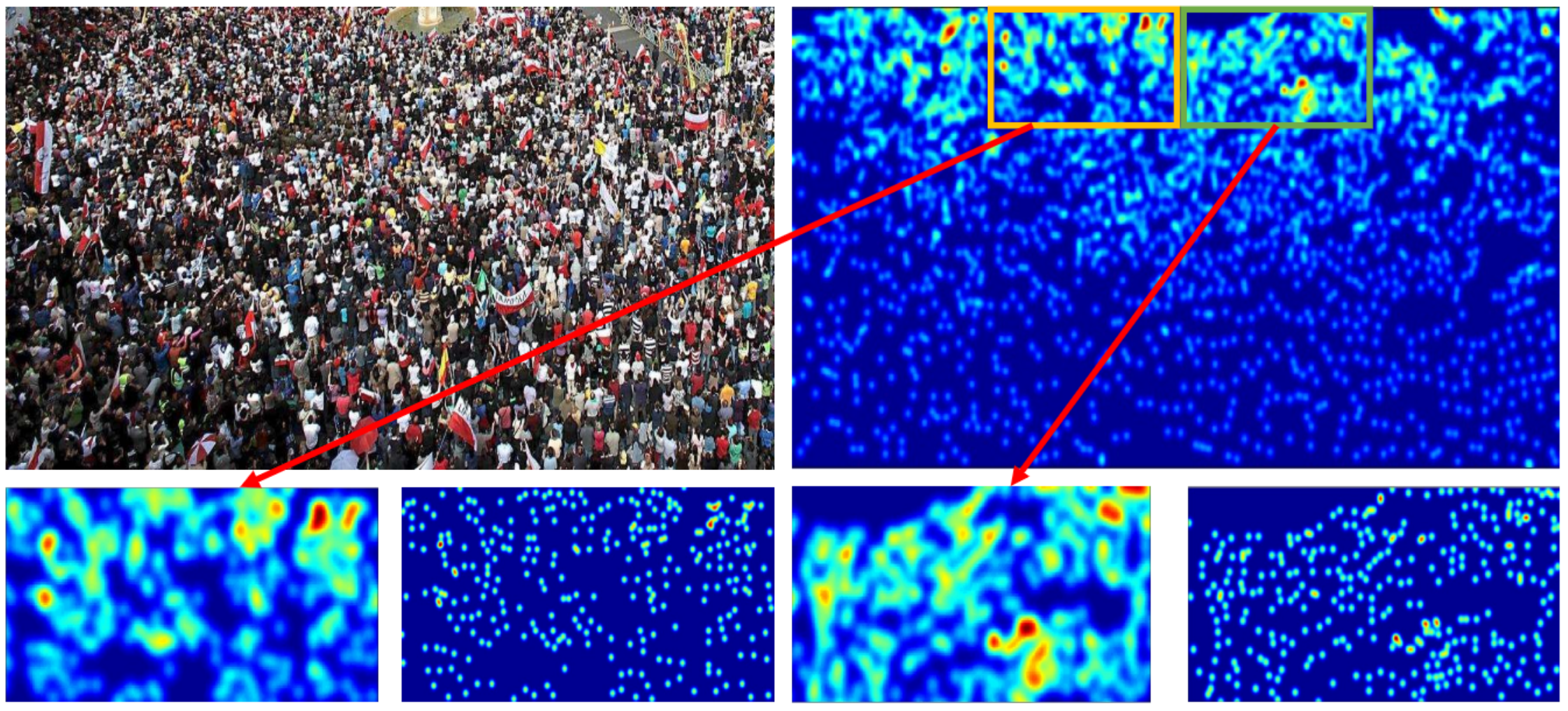}
\caption{An example of ground-truth transformation for density map re-computation by enlarging the distance between blobs while keeping the original peaks, alleviating the density pattern shift between sparse and dense regions. 
}
\label{fig:gt_trans}
\end{figure}

To guide the density map re-prediction on the selected regions, we also adjust the ground-truth density map for each region accordingly. For each selected region $R_i$, instead of directly scaling the ground-truth density map in the same way as feature map scaling, we first scale the binary head location map, and then recompute the ground-truth density map $D'_i$ for $R_i$ by ${D'_i}(p)=\sum_{m=1}^{P_i}{\delta(p -r_i*p_m)}*G_{\sigma_m}(p)$, where $P_i$ is the number of people in $R_i$. As shown in Fig.~\ref{fig:gt_trans}, such ground-truth transformation for density map re-computation reduces the density pattern gap between sparse regions and dense regions, facilitating the density map re-prediction.

The main issue of this density map re-prediction by learning to scale dense regions is to compute appropriate scale ratios for the selected dense regions. Yet, there is no explicit target scale suggesting how much region $R_i$ should be zoomed ideally. We would like to have the estimated average density $\overline{D}_i$ approaching the ground-truth average density on the $i$-th region. The relative density degree of region $R_i$ could be well reflected by $d_i = \overline{D}_{i}/r_{i}^2.$
Assuming that we make the value of $d_i$ for each region close to one of the multiple learnable centers, then we centralize all the selected regions to multiple similar density levels, alleviating the large density pattern shift and thus improving the prediction accuracy. This motivates us to resort to center loss on $d_i$ with multipolar centers. Put it simply, we attempt to centralize all the selected regions into $C$ centers following their average density $\overline{D}$ acting as the unsupervised clustering.  

Specifically, we extend the center loss to a multipolor center loss (MPCL) to handle different density levels. We first initialize the $C$ centers with increasing random values for more and more dense regions. Then for each center $\overline{d}_c$, we follow the standard process of using center loss and update the center for $(t+1)$-th iteration as
\begin{equation}
\Delta{\overline{d_c}^t} = \frac{\sum_{i=1}^{n_c}{(\overline{d_c}^t-\frac{\overline{D}_i^c}{{r_i^c \times r_i^c}})}}{1+n_c}, \, \overline{d_c}^{t+1} = \overline{d_c}^t- \alpha \cdot \Delta{\overline{d_c}^t},
\end{equation}
where $n_c$ refers to the number of regions, $\overline{D}_i^c$ refers to average density map, $r_i^c$ refers to scaling ratio for $i$-th region, and $\alpha$ denotes the learning rate for updating each center, respectively. The $\overline{D}_i^c$ will be centralized to the $c$-th center in an image.
During each iteration, we use the selected $N = \sum_{c=1}^C{n_c}$ dense regions to compute the center loss $L_c$ with multiple centers and update network parameters as well as the centers. The supervision on $r$ using multipolar center loss is the key to bring all the selected regions to multiple similar density levels, leading to robust density estimations.



\subsection{Training objective}
\label{lossfunc}
The whole network is end-to-end trainable, which involves three loss functions: 1) L2 loss for initial prediction of density map $L_D$ given by $L_D = \left\|D - \hat{D}\right\|_2$; 2) 
L2 loss for density map re-prediction on $N = \sum_{c=1}^C{n_c}$ selected regions $L_r$ given by $L_r = \sum_{i = 1}^N\left\| D'_i - \hat{D'}_i\right\|_2$, where $\hat{D}'_i$ denotes the re-predicted density map on the scaled selected region $R_i$; 3) Multipolar center loss at relative density level $d$ for the selected regions $L_c$ computed by
\begin{equation}
    L_c \, = \, \sum_{c=1}^{C}{\sum_{i = 1}^{n_c} \left\| \frac{\overline{D}_i^c}{r_i^c \times r_i^c} - \overline{d}_c\right\|_2}.
\end{equation}
The final loss function $L$ for the whole network is the combination of the above three losses given by
\begin{equation}
\label{lossequa}
L \, = \,  L_D + \lambda_1 \times L_r +\lambda_2 \times L_c,
\end{equation}
where $\lambda_1$ and $\lambda_2$ are two hyperparameters. Note that we optimize the loss function $L$ in Eq.~\eqref{lossequa} to update not only the overall network parameters but also the centers $\{\overline{d}_c\}$.


\begin{table*}
\footnotesize
\centering
\small
\begin{tabular}{ |l|*2{p{1cm}<{\centering}}|*2{p{1cm}<{\centering}}|*2{p{1cm}<{\centering}}|*2{p{1cm}<{\centering}}| }
 \hline
 {\multirow{2}{*}{Method}} & \multicolumn{2}{c|}{ShanghaiTech Part A} & \multicolumn{2}{c|}{ShanghaiTech Part B} & \multicolumn{2}{c|}{UCF\_CC\_50} &\multicolumn{2}{c|}{UCF-QNRF}  \\
 \cline{2-9}
 {} & MAE & MSE & MAE & MSE&MAE&MSE&MAE&MSE \\
 \hline
 \hline
 MCNN~\cite{zhang2016single} & 110.2 & 173.2 & 26.4 & 41.3&377.6&509.1 &277 & -   \\
 CMTL~~\cite{sindagi2017cnn} & 101.3 &152.4 &20.0&31.1& 322.8&397.9&252&514 \\
 Switch-CNN~\cite{sam2017switching} & 90.4&135.0 & 21.6  & 33.4 &318.1 &439.2 &228 &445 \\
 CP-CNN~\cite{sindagi2017generating} & 73.6 & 112.0  & 20.1 & 30.1&298.8&320.9& -& -  \\
 ACSCP~\cite{shen2018crowd} & 75.7 & 102.7  & 17.2 & 27.4&291.0&404.6& -&- \\
 L2R~\cite{liu2018leveraging} & 73.6 & 112.0 & 13.7 & 21.4 &279.6&388.9&- & -  \\
D-ConvNet-v1~\cite{shi2018crowd}&73.5 &112.3 &18.7&26.0&288.4&404.7&-&-\\
 CSRNet~\cite{li2018csrnet} & 68.2 & 115.0  & 10.6 & 16.0&266.1&397.5&-&-\\
 ic-CNN~\cite{ranjan2018iterative} & 69.8 & 117.3  & 10.7 & 16.0&260.9&365.5&-&- \\
 SANet~\cite{cao2018scale} & 67.0 &104.5  & 8.4 & 13.6&258.4&334.9&-&-\\
 CL~\cite{idrees2018composition}&-&-&-&-&-&-&132&191\\
 \hline
 \hline
 VGG16 (\textbf{ours}) & 72.9 & 114.5 & 12.1 &20.5&225.4 &372.5& 120.6&205.2 \\
 SPN (\textbf{ours}) & 70.0 & 106.3 & 9.1 & 14.6&204.7&340.4&110.3&184.6\\
 SPN+L2SM (\textbf{ours})& \textbf{64.2}&\textbf{98.4} &\textbf{7.2} &\textbf{11.1}&\textbf{188.4}&\textbf{315.3}& \textbf{104.7}&\textbf{173.6} \\
 \hline
\end{tabular}
\caption{Quantitative comparison of the proposed method with state-of-the-art methods on three benchmark0 datasets.}
\label{table:Shanghaitech}
\end{table*}


\begin{table*}
\centering
\small
\begin{tabular}{ |l|c|c|c|c|c|c|c|c| }
\hline
{\multirow{2}{*}{Method}}&\multirow{2}{*}{SPN}&\multicolumn{1}{c|}{L2SM (G=3)}&\multicolumn{1}{c|}{L2SM (G=4)}&\multicolumn{5}{c|}{L2SM/S2AD (G=5)} \\
\cline{3-9}
{} &{} & $C=2$ & $C=2$ & $C=1$& $C=2$ & $C=3$ & $C=4$ & $C=5$\\
\hline
\hline
MAE& 70.0  & 65.1 &66.1  & 67.2/68.9 & 65.4/68.1& 64.2/67.0 & 67.1/69.2& 69.8/73.6\\
MSE& 106.3 & 100.4  &103.5   &102.3/110.3 &100.7/107.3 &98.4/105.4 & 101.6/108.7& 104.5/113.5\\
Cost time (s)&0.524 &0.576 &0.569  &0.539/0.540&0.550/0.551 & 0.565/0.563 & 0.583/0.580 & 0.592/0.587 \\
\hline
\end{tabular}

\caption{Ablation study on different settings of dense region selection, number of centers $C$, and different ways of learning to scale. L2SM denotes the proposed learning to scale module and S2AD denotes that we directly scale the selected regions to the average density.}
\label{table:center_number}
\end{table*}

\begin{table}
\small
\centering
\begin{tabular}{ |l|c|c| }
\hline
$K \times K$ setting &MAE&MSE\\
\hline
\hline
$2 \times 2$&68.0&107.1 \\
$4 \times 4$&67.2&106.3 \\
$6 \times 6$&67.9&106.9 \\
$8 \times 8$&68.5&109.1\\
\hline
\end{tabular}
\caption{Ablation study on $K \times K$ image domain divisions for selecting dense region to re-predict under one center setting.}
\label{table:division}
\end{table}

\begin{table*}
\centering
\small
\begin{tabular}{ |l|c|c|c|c|c|c|c|c|c|c| }
\hline
{\multirow{2}{*}{Method}}&\multicolumn{2}{c|}{Part A$\rightarrow$Part B}&\multicolumn{2}{c|}{Part B$\rightarrow$Part A}&\multicolumn{2}{c|}{Part A$\rightarrow$UCF\_CC\_50} &\multicolumn{2}{c|}{UCF-QNRF$\rightarrow$Part A}&\multicolumn{2}{c|}{Part A$\rightarrow$UCF-QNRF}\\
\cline{2-11}
{} & MAE & MSE & MAE & MSE&MAE&MSE&MAE&MSE&MAE&MSE\\
\hline
\hline
MCNN~\cite{zhang2016single} & 85.2 & 142.3 & 221.4&357.8 & 397.7 &624.1&- &-&-&-\\
D-ConvNet-v1~\cite{shi2018crowd}& 49.1 & 99.2 & 140.4&226.1&364&545.8& -&-&-&-\\
L2R~\cite{liu2018leveraging}&-&-&-&-&337.6&434.3 &- &-&-&- \\
\hline
\hline
SPN (\textbf{ours}) & 23.8 &44.2  & 131.2&219.3& 368.3 &588.4 &87.9 &126.3&236.3&428.4\\
SPN+L2SM (\textbf{ours}) & \textbf{21.2}&\textbf{38.7} &\textbf{126.8} &\textbf{203.9} &\textbf{332.4} & \textbf{425.0} &\textbf{73.4}&\textbf{119.4}&\textbf{227.2}&\textbf{405.2}\\
\hline
\end{tabular}
\caption{Cross dataset experiments on ShanghaiTech, UCF\_CC\_50, and UCF-QNRF dataset for assessing the transferability of different methods.}
\label{table:transfer}
\end{table*}

\section{Experiments}
\label{sec:experiments}


\subsection{Datasets and Evaluation Metrics}

We conduct experiments on three widely adopted benchmark datasets including ShanghaiTech~\cite{zhang2016single}, UCF\_CC\_50~\cite{idrees2013multi}, and UCF-QNRF~\cite{idrees2018composition} to demonstrate the effectiveness of the proposed method. These three datasets and the adopted evaluation metrics are shortly described in the following. 


\vspace{1ex}\noindent\textbf{ShanghaiTech Dataset.}~The ShanghaiTech crowd counting dataset~\cite{zhang2016single} consists of 1198 annotated images divided into two parts. Part A contains 482 images which are randomly crawled from the Internet. Part B includes 716 images which are taken from the busy streets of metropolitan area in Shanghai city. 

\vspace{1ex}\noindent\textbf{UCF\_CC\_50 Dataset.}~This dataset is a collection of 50 images of very crowd scenes~\cite{idrees2013multi}. There the number of people varies from 94 to 4543 in images. Following classical benchmarks on this dataset, we use 5-fold cross-validation to evaluate the performance of our method.


\vspace{1ex}\noindent\textbf{UCF-QNRF.}~UCF-QNRF dataset is the recent dataset~\cite{idrees2018composition} containing 1535 images. The number of people in an image varies from 49 to 12865, making this dataset feature huge density variance. Furthermore, the images in this dataset also have very huge resolution variance (\textit{e.g.}, ranging from $400 \times 300$ to $9000 \times 6000$).

\vspace{1ex}\noindent\textbf{Evaluation metrics.}~We employ two standard metrics,~\textit{i.e.},~Mean Absolute Error (MAE) and Mean Squared Error (MSE). MAE and MSE are defined as
\begin{equation}
\text{MAE}=\frac{1}{M}\sum_{i = 1}^M{|c_i-\hat{c_i}|},
\text{MSE}=\sqrt{\frac{1}{M}\sum_{i = 1}^M{{(c_i-\hat{c_i})}^2}},
\end{equation}
where $c_i$ (\textit{resp.} $\hat{c_i}$) represents the ground-truth (\textit{resp.} estimated) number of pedestrians in the $i$-th image, and $M$ is the total number of testing images.

\subsection{Implementation Details}

We follow the setting in~\cite{li2018csrnet} to generate the ground-truth density map. For a given dataset, we first evenly divide all the images in a dataset into $G$ groups of regions with increasing densities, and then attempt to centralize the top $C$ densest groups of regions to $C$ similar density levels ({\textit i.e.}, $C$ centers involved in the center loss), respectively. In the following, without explicitly specifying, $G$ is set to 5, and $C$ is set to 3 for all the used datasets except for UCF\_CC\_50 dataset. Since images from UCF\_CC\_50 dataset consist of crowded people over the whole image domain, we centralize all regions to $C = 5$ similar density levels. Without explicitly specified, the hyperparameter $K$ involved in dividing each image into $K \times K$ regions is set to 4.
 
The loss function described in Eq.~\eqref{lossequa} is used for the model training. We set $\lambda_1$ to 1 and discuss the impact of $\lambda_2$ in Eq.~\eqref{lossequa} in the following. We use Adam~\cite{kingma2014adam} optimizer to optimize the whole architecture with the learning rate initialized to 1e-4. When training on the UCF-QNRF dataset containing images of very high resolutions (\textit{e.g.},~$9000 \times 6000$), we first down-sample the image of which resolution is larger than 1080p to $1920 \times 1080$. Then we divide each image into $2 \times 2$ and combine them into a tensor with batch size equal to 4. When training on the other datasets, we directly input the whole image to our network. 

During inference, we first generate an initial density map $\hat{D}$ for the whole input image, and then select dense regions from $K \times K$ divisions based on the average initial density $\overline{D}_i$ on each region $R_i$. If $\overline{D}_i$ is larger than a predefined value for selecting the top $C$ groups of regions in training, we replace the initial density map prediction with scaled re-prediction for each selected dense region $R_i$.


The proposed method is implemented in Pytorch~\cite{paszke2017automatic}. All experiments are carried out on a workstation with an Intel Xeon 16-core CPU (3.5GHz), 64GB RAM, and a single Titan Xp GPU.


\begin{figure}[t]
\centering
\includegraphics[width=0.8 \linewidth]{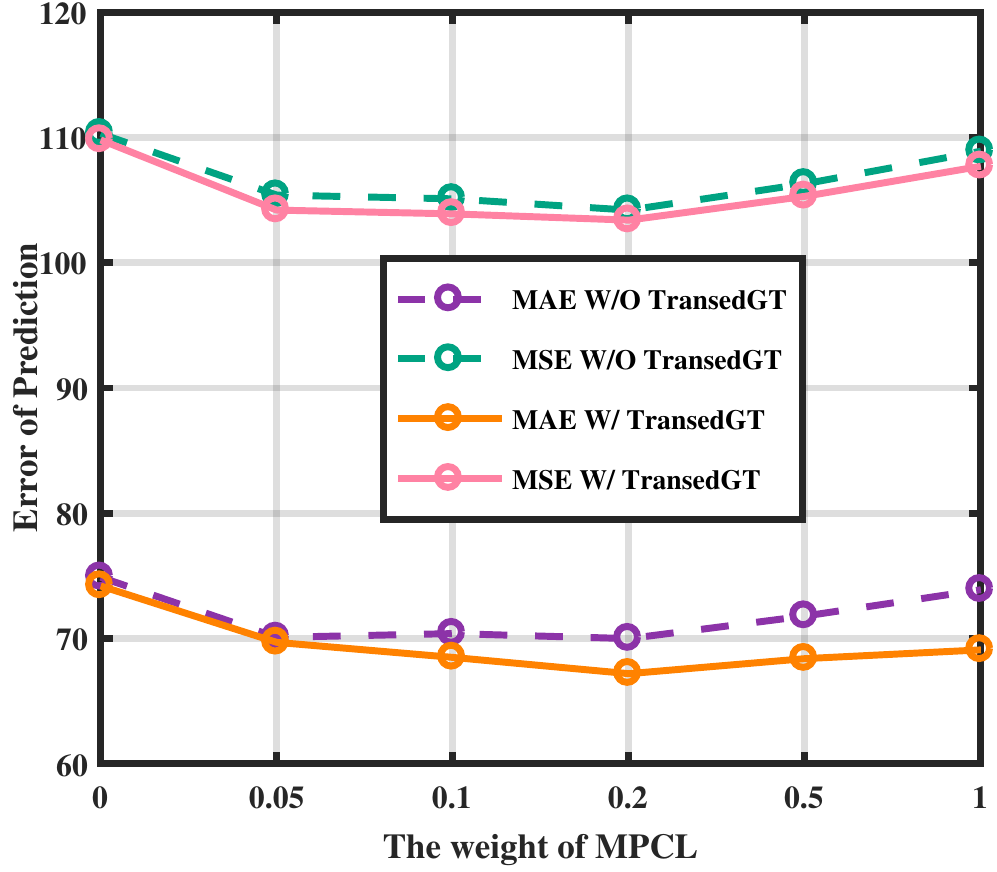}
\centering
  \caption{Ablation study on the effect of weight of the center loss under one center and on whether using ground-truth transformation when scaling for re-prediction. W/ TransedGT means ground-truth transformation is used while W/O TransedGT means it is not used.}
\label{fig:center_weight}
\end{figure}

\begin{figure*}[ht]
\centering
\includegraphics[width=0.8\paperwidth]{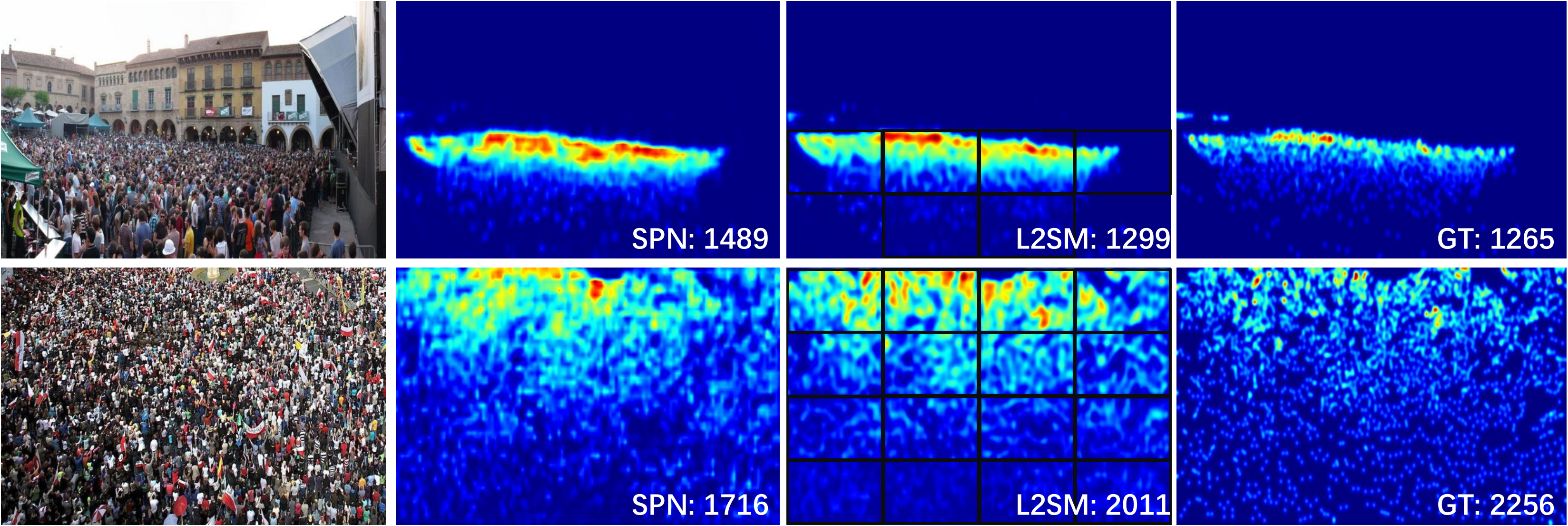}
\centering
\caption{Qualitative visualization of predicted density map on two examples. From left to right: original image, prediction given by SPN, re-predicted density map with L2SM on selected regions (englobed by black boxes), and ground-truth density map.}
\label{fig:example_vis}
\end{figure*}

\subsection{Experimental Comparisons}
 The proposed method outperforms all the other competing methods on all the benchmarks. The quantitative comparison with the state-of-the-art methods on these three datasets is presented in Table~\ref{table:Shanghaitech}.  

\vspace{1ex}\noindent\textbf{ShanghaiTech.}~Our work outperforms SANet~\cite{cao2018scale}, the state-of-the-art method, by 2.8 in MAE and 6.1 in MSE on ShanghaiTech Part A and 1.2 in MAE and 2.5 in MSE on ShanghaiTech Part B. 
It is shown in Table~\ref{table:Shanghaitech} that L2SM improves the performance of our SPN baseline by 5.8 in MAE and 7.9 in MSE on ShanghaiTech Part A, and 1.9 in MAE and 3.5 in MSE on ShanghaiTech Part B. In fact, ShanghaiTech Part A contains images that are more crowded than ShanghaiTech Part B, and the density distribution of ShanghaiTech Part A varies more significantly than that of ShanghaiTech Part B. This may explain that the improvement of the proposed L2SM on ShanghaiTech Part A is more significant than that on ShanghaiTech Part B. 

\noindent\textbf{UCF\_CC\_50.}
We then compare the proposed method with other related methods on UCF\_CC\_50 dataset. To the best of our knowledge, UCF\_CC\_50 dataset is currently the densest dataset publicly available for crowd counting. The proposed method achieves significant improvement over state-of-the-art methods. Precisely, the proposed method decreases the MAE from 258.4 to 188.4, and MAE from 334.9 to 315.3 for SANet~\cite{cao2018scale}. 

\noindent\textbf{UCF-QNRF.}
We also conduct experiments on UCF-QNRF dataset containing images of significantly mulitiple density distributions and resolutions. By limiting the maximal image size to $1920 \times 1080$, our VGG16 baseline already achieves state-of-the-art performance. The proposed SPN brings an improvement of 10.3 in MAE and 20.6 in MSE compared with VGG16 baseline. The proposed L2SM further boosts the performance by 5.6 in MAE and 11.0 in MSE.



\subsection{Ablation Study} \label{section:abltion study}
The ablation studies are mainly conducted on the ShanghaiTech part A dataset, as it is a moderate dataset, neither too dense nor too sparse, and covers a diverse number of people heads.

\vspace{1ex}\noindent\textbf{Effectiveness of different learning to scale settings.}~For the learning to scale process, we first evenly divide the images in a whole dataset into $G$ groups of regions with increasing density, and then attempt to centralize the densest $C$ groups of regions to $C$ similar density levels. As shown in Table~\ref{table:center_number}, the number of groups $G$ and the number of centers $C$ are important for accurate counting. For a fixed number of groups ({\textit e.g.}, $G = 5$), centralizing more and more regions leads to slightly improved counting results. Yet, when we attempt to centralize every image region, we also re-predict the density map for very sparse or background regions, bringing more background noise and thus yielding slightly decreased performance. A relative finer group divisions with a proper number of centers performs slightly better. As shown in Table~\ref{table:center_number}, the proposed L2SM with multipolar center loss performs much better than directly scaling the regions to the average density (S2AD) in each group.



\vspace{1ex}\noindent\textbf{Time overhead.}~To analyze the time overhead of the proposed L2SM, we conduct experiments under seven different settings (see Table~\ref{table:center_number}). The time overhead analysis is achieved by calculating the average inference time on the whole ShanghaiTech Part A test set. The batch size is set to 1 and only 1 Titan-X GPU is used during inference. The average time overhead of SPN is about 0.524s per image. When we increase the number of centers and the number of regions to be re-predicted, the runtime slightly increases. When using 5 centers and re-predict all the $K \times K$ regions, the proposed L2SM increases the runtime by 0.068s per image, which is negligible compared with the whole runtime. 


\vspace{1ex}\noindent\textbf{Effectiveness of the weight of MPCL.}~We study the effectiveness of center loss on ShanghaiTech Part A using one center by changing its weight $\lambda_2$ in Eq.~\eqref{lossequa}. Note that when the weight $\lambda_2$ is set to 0, the center loss is not used, which means that the scale ratio $r$ is learned automatically without any specific supervision. As shown in Fig.~\ref{fig:center_weight}, The use of center loss which brings regions of significantly multiple density distributions to similar density levels plays an important role in improving the counting accuracy. It is also noteworthy that the performance improvement is rather stable for a wide range of weight of the center loss. 

\vspace{1ex}\noindent\textbf{Effectiveness of the ground-truth transformation.}~We also study the effect of ground-truth transformation involved in scale to re-predict process. As shown in Fig.~\ref{fig:center_weight},
the ground-truth transformation by enlarging the distance between crowded heads is more accurate than straightforwardly scale the ground-truth density map.
It is not surprised to understand that enlarging the distance between crowded heads results in regular Gaussian density blobs for dense regions, which reduces the density pattern shift thus facilitates the density map prediction.


\vspace{1ex}\noindent\textbf{Effectiveness of the division.}~We also conduct experiments by varying the $K \times K$ image domain divisions. As shown in 
Table~\ref{table:division}. The performance is rather stable across different image domain division. 


\subsection{Evaluation of Transferability}

To demonstrate the transferability of the proposed method across datasets, we conduct experiments under cross dataset settings, where the model is trained on the source domain and tested on the target domain.

The cross dataset experimental results are presented in Table~\ref{table:transfer}. We can observe that the proposed method generalizes well to unseen datasets. In particular, the proposed method consistently outperforms D-ConvNet-v1~\cite{shi2018crowd} and MCNN~\cite{zhang2016single} by a large margin. The proposed method also performs slightly better than L2R~\cite{liu2018leveraging} in transferring models trained on ShanghaiTech Part A to UCF\_CC\_50. Yet, the improvement is not as significant as the comparison with~\cite{zhang2016single,shi2018crowd} on transferring between ShanghaiTech Part A and Part B. This is probably because L2R~\cite{liu2018leveraging} also relies on extra data which may somehow help to reduce the gap between the two datasets. As shown in Table~\ref{table:transfer}, the proposed L2SM plays an important role in ensuring the transferability of the proposed method. 
Furthermore, as shown in Table~\ref{table:Shanghaitech} and Table~\ref{table:transfer}, the proposed method under cross-dataset settings performs competitively or even outperforms some methods~\cite{sam2017switching,sindagi2017generating,sindagi2017cnn,zhang2016single} using the proper training set. This also confirms the generalizability of the proposed method.

\section{Conclusion}
In this paper, we propose a Learning to Scale Module (L2SM) to tackle the problem of large density variation for crowd counting. We achieve density centralization by a novel use of multipolar center loss. The L2SM can effectively learn to scale significantly multiple dense regions to multiple similar density levels, making the density estimation on dense regions more robust. Extensive experiments on three challenging datasets demonstrate that the proposed method achieves consistent and significant improvements over the state-of-the-art methods. L2SM also shows the noteworthy generalization ability to unseen datasets with significantly different density distributions, demonstrating the effectiveness of L2SM in real applications.

\section*{Acknowledgement}
This work was supported in part by the National Key Research
and Development Program of China under Grant 2018YFB1004600,
in part by NSFC 61703171, and in part by NSF of Hubei Province of China under Grant 2018CFB199, to Dr. Yongchao Xu by the Young Elite Scientists Sponsorship Program by CAST.

{\small
\bibliographystyle{ieee}
\normalem
\bibliography{egbib}
}

\end{document}